\documentclass[letterpaper, 10 pt, conference]{ieeeconf}  

\IEEEoverridecommandlockouts                              

\usepackage{acro} 
\usepackage{gensymb}
\usepackage{float}
\usepackage{bm}
\usepackage{gensymb}
\usepackage[binary-units,product-units=power]{siunitx}
\usepackage{graphicx}

\usepackage{mathtools}
\usepackage{amsmath}
\usepackage{amssymb}
\usepackage{amsfonts}
\usepackage[mode=buildnew]{standalone}
\usepackage{booktabs} 
\usepackage[caption=false,font=footnotesize]{subfig}
\usepackage{nccmath}
\usepackage[hidelinks]{hyperref}

\usepackage{cleveref}

\usepackage[style=ieee,citestyle=numeric-comp]{biblatex}
\DeclareAcronym{DOF}{
  short = DOF,
  long  = degrees of freedom
}
\DeclareAcronym{GNSS}{
  short = GNSS,
  long  = global navigation satellite system
}

\DeclareAcronym{IMU}{
  short = IMU,
  long  = Inertial Measurement Unit
}
\DeclareAcronym{LiDAR}{
  short = LiDAR,
  long  = Light Detection and Ranging
}
\DeclareAcronym{UXO}{
  short = UXO,
  long  = unexploded ordnance
}

\DeclareAcronym{EKF}{
  short = EKF,
  long  = Extended Kalman Filter
}
\DeclareAcronym{iEKF}{
  short = iEKF,
  long  = iterated Extended Kalman Filter
}

\DeclareAcronym{LIO}{
  short = LIO,
  long  = LiDAR-Inertial Odometry
}

\DeclareAcronym{LO}{
  short = LO,
  long  = LiDAR Odometry
}

\DeclareAcronym{MAV}{
  short = MAV,
  long  = micro aerial vehicle,
  short-indefinite  = an
}

\DeclareAcronym{FOV}{
  short = FOV,
  long  = field of view
}

\DeclareAcronym{SDF}{
  short = SDF,
  long  = signed distance field,
  short-indefinite  = an
}

\DeclareAcronym{NCCR}{
  short = NCCR,
  long  = National Center of Competence in Research,
  short-indefinite = an,
  long-indefinite = a
}

\DeclareAcronym{SNR}{
  short = SNR,
  long  = Signal-to-Noise Ratio,
}
\DeclareMathOperator{\atantwo}{atan2}

\begin{document}
\title{\LARGE \bf
COIN-LIO: Complementary Intensity-Augmented LiDAR Inertial Odometry
\vspace{-3mm}}

\author{Patrick Pfreundschuh, Helen Oleynikova, Cesar Cadena, Roland Siegwart, and Olov Andersson%
\vspace{-4.5em}
\thanks{Authors are with Autonomous Systems Lab, ETH Zurich, e-mail: \texttt{\small patripfr@ethz.ch}. This work was supported by Swiss National Science Foundation’s NCCR DFab P3.
\vspace{-0.5mm}}%
}

\maketitle

\begin{abstract}
We present COIN-LIO, a LiDAR Inertial Odometry pipeline that tightly couples information from LiDAR intensity with geometry-based point cloud registration. The focus of our work is to improve the robustness of LiDAR-inertial odometry in geometrically degenerate scenarios, like tunnels or flat fields. We project LiDAR intensity returns into an image, and present a novel image processing pipeline that produces filtered images with improved brightness consistency within the image as well as across different scenes. We effectively leverage intensity as an additional modality, using our new feature selection scheme that detects uninformative directions in the point cloud registration and explicitly selects patches with complementary image information. Photometric error minimization in the image patches is then fused with inertial measurements and point-to-plane registration in an iterated Extended Kalman Filter. The proposed approach improves accuracy and robustness on a public dataset. We additionally publish a new dataset, that captures five real-world environments in challenging, geometrically degenerate scenes. By using the additional photometric information, our approach shows drastically improved robustness against geometric degeneracy in environments where all compared baseline approaches fail.
\end{abstract}

\section{INTRODUCTION}
Recent advances in 3D \ac{LiDAR} have decreased both the size and price of these sensors, enabling them to be used by a wider range of robots. At the same time, new LiDAR-based state estimation approaches such as FAST-LIO2~\cite{fastlio2} have increased the accuracy and robustness while decreasing the computational cost, making 3D LiDAR one of the most popular choices for mobile robot sensors.
However, even these \ac{LIO} approaches struggle in geometrically degenerate environments, such as tunnels and flat fields.

\begin{figure}
    \centering
\includegraphics[width=\linewidth]{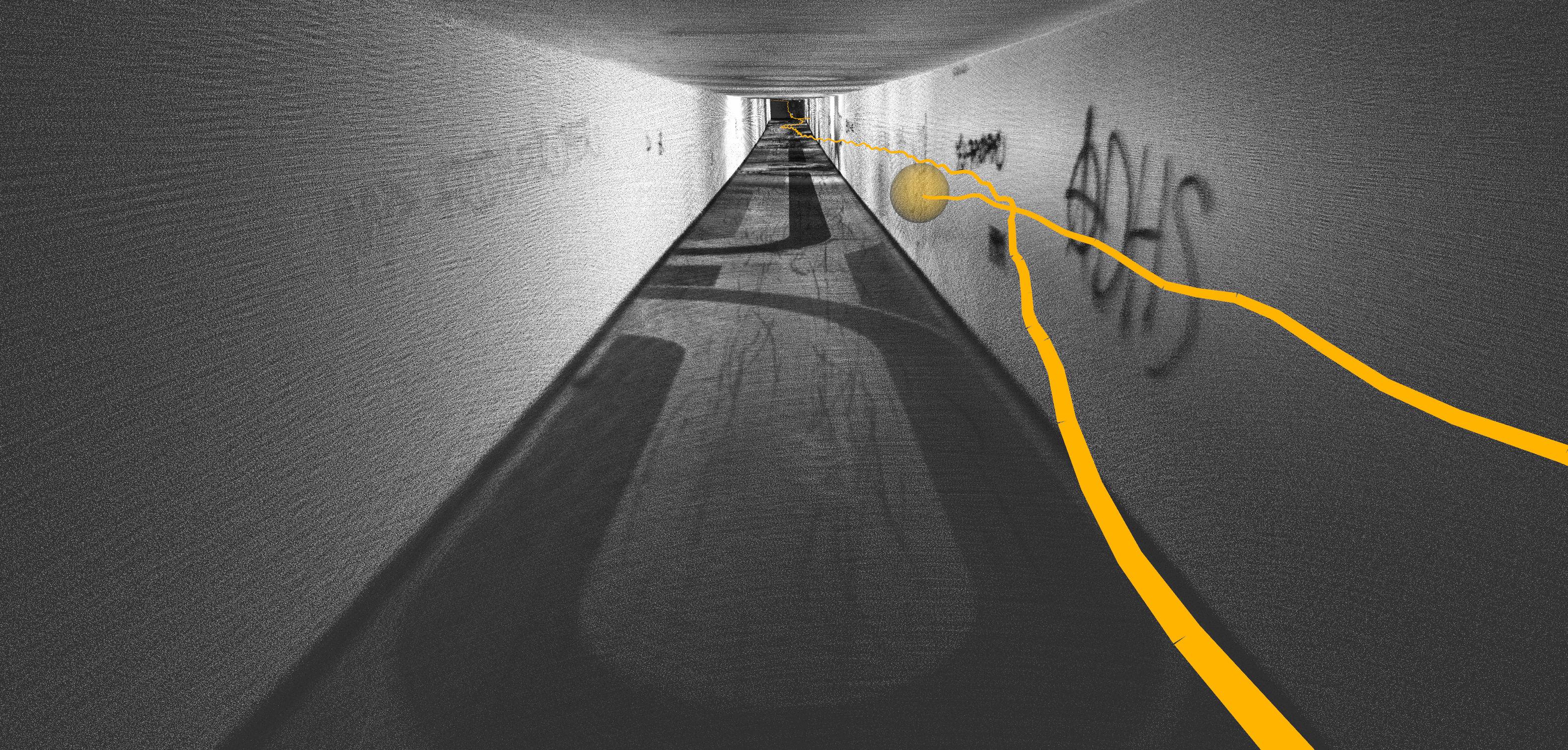}
\adjustbox{trim={0} {.5\height} {0} {0},clip}%
  {\includegraphics[width=\linewidth]{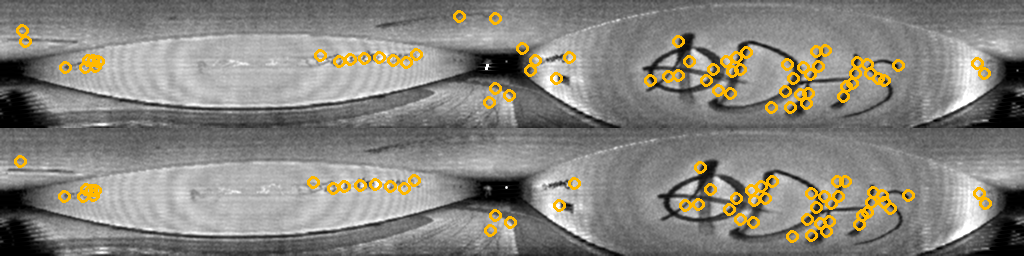}}
\includegraphics[width=\linewidth]{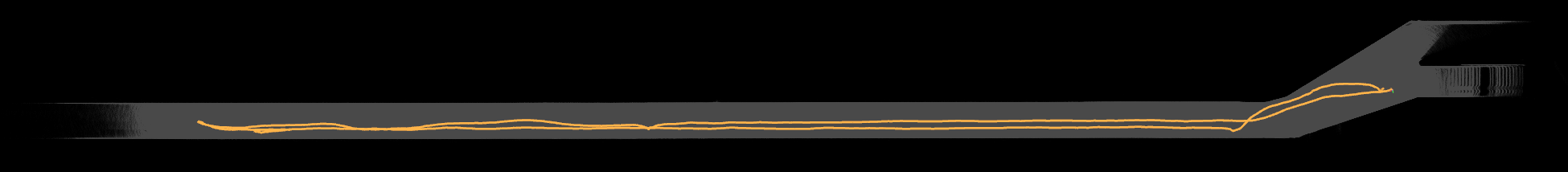}
    \vspace{-6mm}
    \caption{\textit{Top}: Accumulated point cloud colorized by intensity and trajectory (orange) resulting from COIN-LIO. Our approach achieves accurate odometry despite geometric degeneracy along the tunnel, resulting in clearly visible correct ground and wall markings. \textit{Mid}: Filtered intensity with tracked features (orange).
    \textit{Bottom}: Top view of the resulting point cloud (gray) and trajectory (orange) from the tunnel.}
    \label{fig:undulated_heat}
    \vspace{-8mm}
\end{figure}

In most geometrically uninformative scenes, the texture of the environment still offers some visual information.
While several works~\cite{vloam, licfusion2, vilens, r3live} fuse camera information with LiDAR to take advantage of this complementary data, this requires additional sensors, accurate extrinsic calibration, and time synchronization. Cameras will also not work in the absence of ambient light, which limits their applicability. 
However, in addition to range measurements, LiDARs provide the measured signal strength of each reflected point (intensity). For modern mechanically rotating multi-layer LiDARs, this signal can be projected into a dense image, which allows the LiDAR to operate as an active camera without external illumination. Images and point clouds are time-synchronized and the extrinsics are known, which drastically simplifies their use compared to a combination of LiDAR with cameras.

These intensity images contain texture information about the environment, which can be used for pose estimation.

Compared to camera images, intensity images suffer from poor Signal-to-Noise Ratio, lower resolution, strong rolling shutter effects, and a different projection model from traditional pinhole cameras, making it difficult to directly apply existing visual odometry methods. 
While several works have used intensity to improve pose estimation~\cite{mdslam, ri-lio, realtimeint}, they perform no filtering to improve image quality and do not combine the information from geometry and intensity in a complementary way, which limits performance in geometrically degenerate scenes as shown in our experiments. 

To this end, we present COmplementary INtensity-Augmented LIO, a robust, real-time LIO framework, that couples geometric registration with photometric error minimization for increased robustness. We improve upon related work by introducing a filtering method to increase brightness consistency in the intensity image and a feature selection scheme that adds features with complementary information to the degenerate geometry of the scene. 
This feature selection is important as parts of the scene (like edges of a tunnel) are often also visually uninformative along the geometrically degenerate direction.
The complementary information vastly increases the robustness of the combined method in geometrically degenerate scenarios while keeping or improving performance in geometry-rich scenes.

We found a lack of 3D LiDAR datasets focusing on scenarios with degenerate geometry.
To this end, we created the ENWIDE dataset, which captures five real-world ENvironments WIth large sections of DEgenerate geometry and recorded ground truth positions from a high-accuracy laser scanner.
We hope that by providing this data to the community along with our open-sourced code implementation\footnote{\url{https://github.com/ethz-asl/coin-lio}}, we can fuel further advances in robust LiDAR-inertial odometry.

The main contributions of our work can be stated as follows: \textit{(1)} we present a LiDAR-intensity image processing pipeline as well as a geometrically complementary feature selection scheme that enables detection and tracking of salient features with complementary information to the geometry-based measurements, \textit{(2)}  we show that our approach effectively leverages LiDAR intensity to improve robustness and performance of \ac{LIO} in geometrically degenerate scenes, \textit{(3)} we provide a real-world dataset, ENWIDE\footnote{\url{https://projects.asl.ethz.ch/datasets/enwide}}, that contains ten sequences in five scenes of diverse geometrically degenerate environments, with accurate position ground truth.

We present our contributions in a combined system with geometry-based \ac{LIO}, based on FAST-LIO2~\cite{fastlio2}, and show superior performance on a standard dataset and ENWIDE, over geometry-only and geometry-and-intensity-based methods.

\section{Related Work}
\vspace{-0.5mm}
\subsection{LiDAR (Intertial) Odometry}
\vspace{-1mm}
Common LiDAR-based odometry approaches are based on registration of a measured point cloud against a map that is built during operation. For many years, the standard approach for \ac{LO} was LOAM~\cite{loam} which extracts points on edges and planes for registration. This works well in structure-rich environments, but edge and plane points are often not expressive enough to perform robustly in geometrically challenging scenarios. KISS-ICP~\cite{kiss-icp} avoids feature selection and directly registers a voxel-downsampled point cloud with point-to-point ICP which showed improved performance in unstructured environments. X-ICP~\cite{xicp} explicitly detects degenerate directions in the registration but relies on an auxiliary state estimate. The use of inertial measurements in LIO approaches has shown a large increase in robustness, as it helps to remove ego-motion distortion from the point cloud and provides an initial guess for the registration. 
LIO-SAM~\cite{lio-sam} fuses \ac{IMU} measurements in a factor-graph~\cite{isam2, factor_graphs} with edge and plane feature matching against submaps.
FAST-LIO~\cite{fastlio1} presents an efficient formulation of the Kalman Filter update that enables the alignment of every scan against the continuously built map in real-time. The authors switch from feature matching to raw points with point-to-plane ICP in its successor~\cite{fastlio2} that achieves state-of-the-art performance.
\vspace{-5mm}

\subsection{Intensity Assisted Odometry}
\vspace{-1mm}
Several approaches use intensity as a similarity metric and integrate it into a weighted ICP~\cite{i-loam, intensity-slam} or use high-intensity points as an additional feature class~\cite{intensityaugmentedlidarinertial, inten-loam, mulls}, but due to their limited map resolution, these approaches cannot capture fine-grained details. 
Early works~\cite{mcmanus2013towards, barfoot2016into, mcmanus2011towards, dong2013lighting} have shown that LiDAR intensities can create lighting invariant images that can be used for visual odometry. However, as they only match intensity image features, these approaches do not leverage the geometric information efficiently. Similarly, the approaches presented in \cite{fastsparse, realtimeint} detect and match image features in the intensity image for registration. However, in geometrically degenerate cases such features are often sparse and most of the geometric information is neglected, resulting in inferior performance. In these approaches, the intensity only influences the point correspondences but does not directly provide a gradient in the optimization. In contrast, in MD-SLAM~\cite{mdslam}, the photometric error of the intensity image is optimized together with a range and normal image. Unlike our work, they do not use the IMU or perform motion undistortion. They also use the entire dense image instead of sparse informative patches. The approach closest to our work is RI-LIO~\cite{ri-lio}; similar to us, they integrate photometric error minimization into the \ac{iEKF}~\cite{he2021kalman} of \cite{fastlio2} but use reflectivity instead of intensity. They randomly downsample the point cloud and project single points into the reflectivity image for the photometric components. However, relevant information is typically not distributed homogeneously in images but concentrated in specific salient regions. Instead of single random pixels at a low resolution, we specifically select geometrically complementary, salient high-resolution patches from a filtered image and continuously assess the feature quality.
This leads to superior performance in difficult geometrically-deficient scenarios compared to existing approaches.
\vspace{-0.5mm}

\section{Method}
\begin{figure*}[t]
    \vspace{-0.5mm}
   \begin{centering} 
    \includegraphics[width=0.9\textwidth]{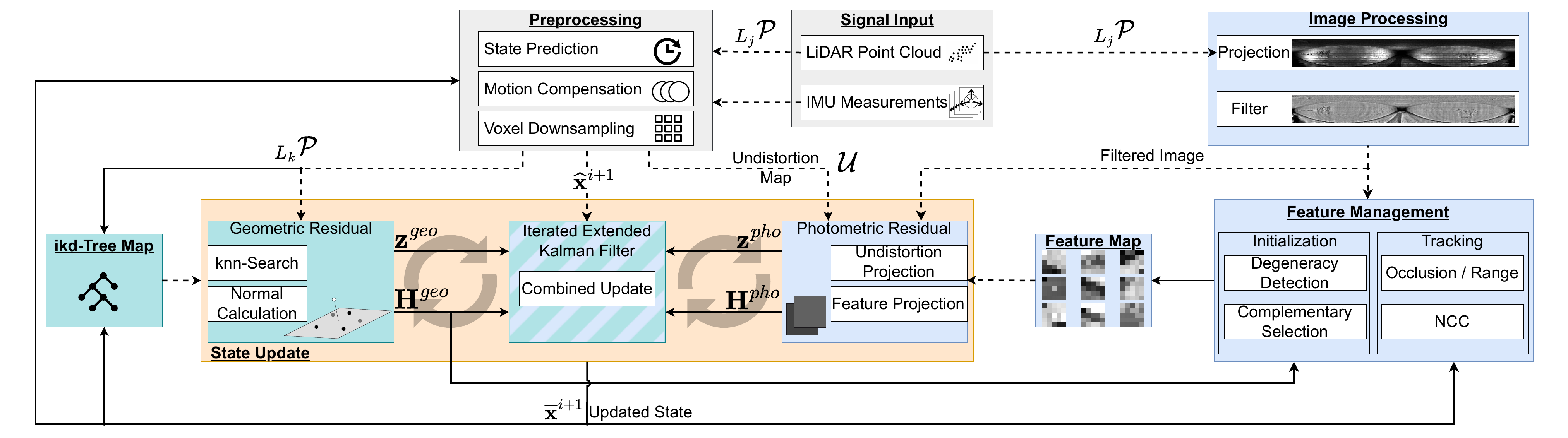}
    \vspace{-4.5mm}
    \caption{System Overview: The input point cloud is used geometrically (green) for map registration and as a projected image (blue) for photometric error minimization. Both residuals are combined in an iterated update (orange). We use the registration Jacobian to find uninformative directions in the geometry and select features with complementary image information (right bottom). Lines indicate information flow \textit{before} (- - -) and \textit{after} (---) the update step.}
    \end{centering} 
    \label{fig:pipeline}
    \vspace{-6mm}
\end{figure*}
COIN-LIO adopts the tightly-coupled \ac{iEKF} presented in FAST-LIO2 for point-to-plane registration and extends it with photometric error minimization. Due to space limitations, we do not review FAST-LIO2~\cite{fastlio1, fastlio2} and focus on the photometric component. We process intensity images from point clouds using a novel filter that improves brightness consistency and reduced sensor artefacts. We specifically select image features that provide information in uninformative directions of the point cloud geometry. The feature management module examines the validity of tracked features and detects occlusions. Finally, we integrate the photometric residual into the Kalman Filter.
\vspace{-2mm}

\subsection{Definitions}
We define a fixed global frame ($\mathit{G}$) at the initial pose of the IMU ($\mathit{I}$). The transformation from LiDAR frame ($\mathit{L}$) to IMU frame is assumed to be known as $\mathbf{T}_{IL} =  (\mathbf{R}_{IL}, {}_{I}\mathbf{p}_{IL})\in \mathit{SE}(3)$.
We define the robot's state as
$\mathbf{x} = [\mathbf{R}_{GI}, {}_{G}\mathbf{p}_{GI}, {}_{G}\mathbf{v}_{I}, \mathbf{b}^{a}, \mathbf{b}^{g}, {}_{G}\mathbf{g}],$
where $\mathbf{R} \in \mathit{SO}(3)$ denotes orientation, $\mathbf{p} \in \mathbb{R}^3$ is the position, $\mathbf{v} \in \mathbb{R}^{3}$ describes linear velocity, and $\mathbf{b}^{a}, \mathbf{b}^{g} \in \mathbb{R}^3$ indicate accelerometer and gyro biases.
The LiDAR frame at $t_j$ is denoted as $L_j$. Each LiDAR scan consists of points recorded during one full revolution $\mathcal{P} = \{{}_{L_j}\mathbf{p}_j, j=1,...,k\}$, with $t_j \leq t_k$. 
\vspace{-1mm}

\subsection{IMU Prediction and point cloud undistortion} \label{undistortion}
We adopt the Kalman Filter prediction step according to FAST-LIO2~\cite{fastlio2} by propagating the state using IMU measurement integration from $t_j$ to $t_k$. Similarly, we calculate the ego-motion compensated, undistorted points at the latest timestamp $t_k$ as: 
${}_{L_k}\mathbf{p}_j = \mathbf{T}_{L_kI_k}\mathbf{T}_{I_kI_j}\mathbf{T}_{I_jL_j}{}_{L_j}\mathbf{p}_j$.

\subsection{Image Projection Model}
\begin{figure}[b!]
    \vspace{-7mm}
    \includegraphics[width=0.6\linewidth]{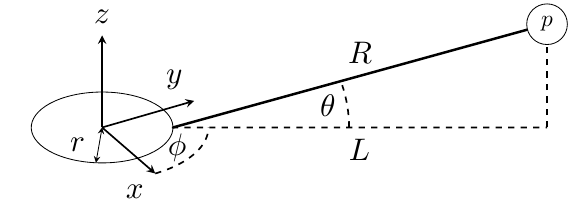}
    \centering
    \vspace{-4mm}
    \caption{Projection model. The offset between LiDAR origin and laser emitter is denoted as $r$. A measured point is depicted on the top right ($p$).}
    \label{fig:project}
\end{figure}
\vspace{-0.5mm}
A point $ {}_{L_j}\mathbf{p}_j = [x_j,y_j,z_j]$ can be projected to image coordinates using a spherical projection:
\vspace{-0.5mm}
\begin{equation} \label{eqn_proj}
\resizebox{0.91\linewidth}{!}{
${}_{C}\mathbf{p}_j  = \Pi({}_{L_j}\mathbf{p}_j) = \begin{bmatrix} \mathit{f}_x\phi + \mathit{c_x}\\ \mathit{f}_y\theta + c_y\end{bmatrix} =  \begin{bmatrix} \frac{-w}{2\pi}\atantwo(\frac{y_j}{x_j}) + \frac{w}{2}\\ \frac{-h}{\Theta_{fov}}\arcsin(\frac{z_j}{R_j}) + \frac{h}{2}\end{bmatrix} = \begin{bmatrix} u_j \\ v_j\end{bmatrix} $} \end{equation} 
with $R =  \sqrt{L^2+z^2}$, $L = \sqrt{x^2+y^2} - r$, as illustrated in Figure \ref{fig:project}. The vertical \ac{FOV} is represented as $\Theta_{fov}$, and $w$ and $h$ denote the horizontal and vertical resolution of the LiDAR. This model assumes a constant elevation angle spacing between subsequent beams.
However, for manufacturing reasons, most LiDARs have an irregular spacing which causes empty pixels in the spherical image~\cite{ri-lio}. While we still use \cref{eqn_proj} to calculate the Jacobian in \cref{eqn:proj_jacobian}, we directly use laser beam and encoder value to create the image. We compensate the horizontal offset similar to \cite{ri-lio}, but use a constant value for all beams. We keep a list of all beam-elevation angles $\Theta_L = \{\theta_1, ..., \theta_h\}$ from the calibration of the LiDAR. When we project a feature point into the image, we calculate $\theta_f$ and find the beams above and below in $\Theta_L$ to interpolate the subpixel coordinate. 
\vspace{-1mm}

\subsection{Image Processing}
\vspace{-0.5mm}
The irregular elevation angle spacing between the beams causes horizontal line artefacts in the intensity image. They are less apparent in structure-rich scenes, but dominate the image in environments with little structure. As they occur at a regular row-frequency we design a finite impulse response filter to remove them. First, we use a highpass filter vertically with a cutoff just below the line frequency. Apart from the lines, the highpass signal also contains relevant image content at this frequency. We therefore apply a lowpass filter horizontally to it, which isolates the lines as relevant image signals appear at a higher horizontal frequency. Finally, we subtract the isolated signal from the intensity image. 
As intensity values depend on the reflectivity of the surface as well as the distance and incidence angle, the intensity is lower in areas that are farther away from the sensor. 

LiDARs such as the Ouster also report compensated reflectivity signals, which is used in \cite{ri-lio}, but the influence of the incidence angle remains. We propose a different approach to achieve consistent brightness throughout the image. 

\begin{figure}[b]
    \vspace{-6mm}
    \includegraphics[width=\linewidth]{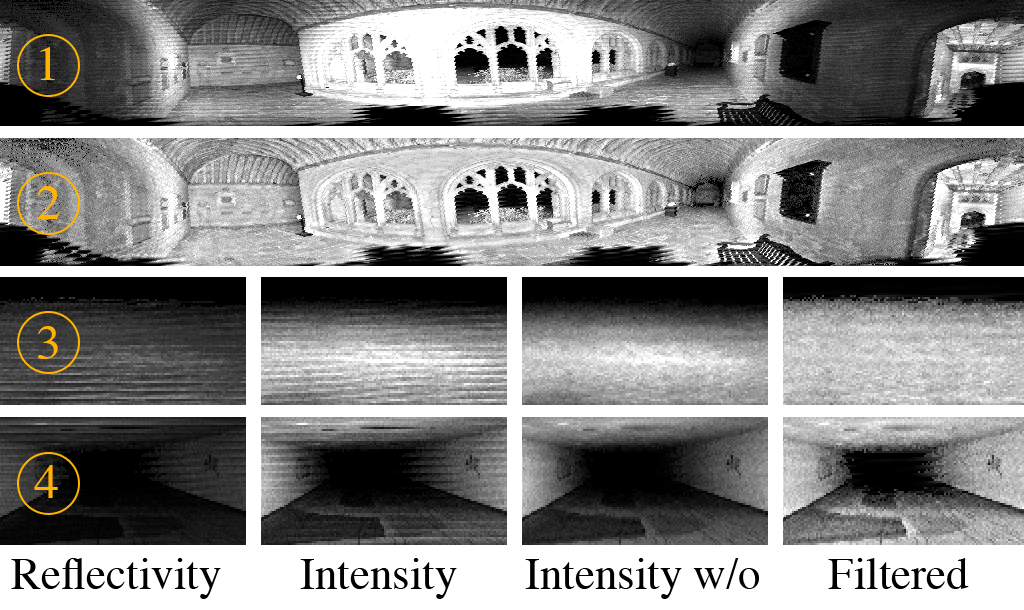}
    \vspace{-9mm}
    \caption{(1): The intensity image is over- (center) and under-exposed (sides). (2): Our filtered image has consistent brightness across the image.
    (3) \& (4): Detail views from a grass field (3) and tunnel (4). The reflectivity image is under-exposed and does not show the ground markings (4). The intensity suffers from strong line artefacts that dominate the texture (3). Our filter removes the line artefacts (Intensity w/o). Our brightness compensation produces consistent exposure and shows details at larger range (ground markings in (4), grass texture in (3)).}
    \label{fig:img_process}
\end{figure}

The brightness level varies smoothly throughout the image, as average distance and incidence angle are typically driven by the global structure of the scene instead of small geometric details. We thus build a brightness map $\mathit{I}_b(u,v)$ by averaging the intensity values in a large window.
To achieve consistent exposure throughout the image, we normalize the pixel values using the brightness map and scale them to values in $[0,255]$ using a constant factor $s_i$:
\vspace{-2mm}
\begin{equation}  
\mathit{I}_F(u,v) = s_i \cdot \frac{\mathit{I}(u,v)} {\mathit{I}_b(u,v) + 1} \vspace{-1.5mm}\end{equation} 
Finally, we smooth the image using a $3 \times 3$ Gaussian kernel to reduce noise. We provide explanatory images in Figure \ref{fig:img_process}. 

\vspace{-1mm}
\subsection{Geometrically Complementary Patch Selection}
\vspace{-1mm}
We select and track $5\times5$ pixel patches which has shown better convergence compared to single pixels~\cite{svo}.
In contrast to prior works that select features randomly~\cite{ri-lio} or based on visual feature detectors~\cite{fastsparse, realtimeint}, we follow an approach inspired by \cite{dso}. 
We select candidate pixels with an image gradient magnitude above a threshold and perform a radius-based non-maximum suppression. This approach does not rely on corner features which is favourable for low-texture images.
Candidate pixels are mostly detected on shape discontinuities in the 3D scene such as edges and corners, or on changes in surface reflectivity, e.g. from ground markings or vegetation. The information from intensity Jacobians of pixels on shape discontinuities often overlaps with the information that is already captured in the point-to-normal registration. We thus aim to select candidates that give additional information to efficiently leverage the multi-modality.

\begin{figure}[]
    \vspace{2mm}
    \includegraphics[width=\linewidth]{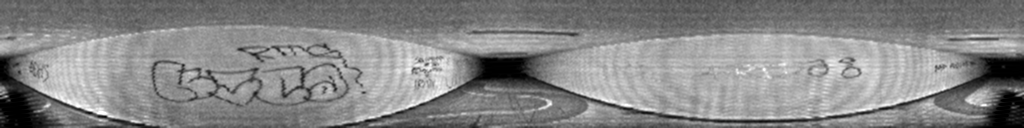}
    \includegraphics[width=\linewidth]{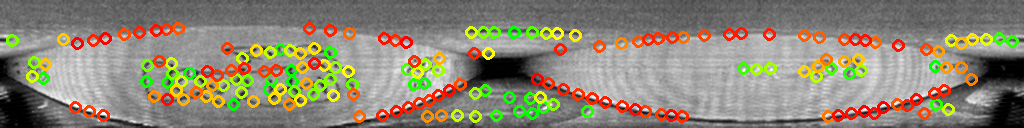}
    \vspace{-7mm}
    \caption{\textit{Top}: Example frame \textit{Bottom}: Features are colored by contribution strength in the uninformative axis along the tunnel (increases from red to green). Uninformative features along the tunnel edges are correctly marked in red, while features with strong gradients along the tunnel show up green.}
    \label{fig:complement}
    \vspace{-6mm}
\end{figure}

To detect uninformative directions in the point cloud registration, we follow the information analysis presented in X-ICP~\cite{xicp}. We calculate the principal components of the Hessian ${\mathbf{H}^{geo}}^T\mathbf{H}^{geo}$ of the point-to-plane terms. A direction is then detected as uninformative if the accumulated filtered contribution is below a threshold. We refer the reader to \cite{xicp} for more details.  
We analyze the translational components and denote the set of uninformative directions $\mathit{V}_t$. If all directions are informative, we insert vectors along the coordinate axes to promote equally distributed gradients.
We calculate the second image moment $M$~\cite{harris} and use its strongest eigenvector $\mathbf{v}_{patch}$ to approximate the patch gradient, which is more stable than pixel gradients. We then calculate how the projected image coordinate changes, if the point is perturbed along a direction using \cref{eqn:jacobian}:
\vspace{-0.5mm}
\begin{equation}\mathbf{d}_{p_i} = \frac{\partial \Pi({}_{L_j}\mathbf{p}_j)}{\partial {}_{L_j}\mathbf{p}_j} \cdot \mathbf{v}_{t,i} \in \mathbb{R}^2, \, \forall \, \mathbf{v}_{t,i} \in V_t \end{equation} 
\vspace{-0.5mm}
We select features where shifting the point along an uninformative 3D direction results in a 2D coordinate shift in an informative image direction. We therefore project the projection gradient $\mathbf{d}_{p_i}$ onto the informative direction $\mathbf{v}_{patch}$ of the patch to calculate its directional contribution $c_i$.
As the magnitude of the projection gradient increases with decreasing range, which would favor the selection of points close to the sensor, we use the normalized gradient instead:
\vspace{-0.5mm}
\begin{equation} c_i =   \frac{\mathbf{d}_{p_i} \cdot \mathbf{v}_{patch}} {||\mathbf{d}_{p_i}||} \end{equation}
\vspace{-0.5mm}
For each direction in $V_t$, we select the patches with the strongest contribution.
We visualize the results in Figure \ref{fig:complement}.

\subsection{Feature Management}
We initialize each point in a patch separately at its global position using the current pose estimate. Different from visual odometry approaches~\cite{svo}, where one position is assigned to the whole patch, this allows us to project each point in the patch separately. Using high-resolution patches we can capture fine-grained details in contrast to prior works~\cite{ri-lio, intensity-slam} which only store a single value per voxel-grid cell. To reduce computational load, we limit the number of tracked patches. After each update step, we assess the feature patch validity. To detect occlusions, we compare the predicted and measured range for each point in the patch and discard all points in it if the difference is above a threshold. We also remove patches below a minimum or above a maximum range. Additionally, we calculate the normalized cross-correlation (NCC) between the tracked and measured patch and remove it if the NCC is below a threshold. We only track features over a maximum amount of frames to reduce error accumulation and to encourage the initialization of new features.  
We avoid overlapping features by enforcing a minimum distance between new and tracked features.

\subsection{Photometric Residual \& Kalman Update} \label{state_estimation}
We minimize photometric errors between tracked and currently observed points. The error is computed by projecting tracked points into the current image and comparing current intensity values to the patch:
\vspace{-1mm}
\begin{equation} \label{eqn_res_phot}
z^{pho} = I_c(\Pi({}_{L_j}\mathbf{p}_f)) - i_f
\vspace{-1mm}
\end{equation} 
As rotating LiDARs record individual points sequentially, the pixels inside the intensity image are measured at different times and different poses. For the projection we therefore need to calculate the position of the tracked point in the distorted LiDAR frame $L_j$:

\begin{equation} \label{eqn_pflj}
{}_{L_j}\mathbf{p}_f =  \mathbf{T}_{L_jI_j} \mathbf{T}_{I_jI_k} \mathbf{T}_{I_kG} {}_{G}\mathbf{p}_f
\end{equation} 
However, this is dependent on $\mathbf{T}_{I_jI_k}$, which in turn depends on the unknown time $t_j$ itself. RI-LIO solves this by using a kNN-search in a kD-tree. However, this is only computationally feasible at a low resolution. We thus propose a projection-based solution.
Given the undistorted point cloud, we can approximate which volumetric slices of the environment were captured at which timestamp. Therefore, we build an undistortion map by projecting the undistorted point cloud into an image and assign each pixel the index of the corresponding point: $\mathcal{U}(\Pi({}_{L_k}\mathbf{p}_j)) = j$.

To find the corresponding index for the feature point, we project it to the undistortion map, which is drastically cheaper than kD-tree-search and thus applicable to the full resolution point cloud. Given the index, we find the respective timestamp and $\mathbf{T}_{I_jI_k}$ to calculate \cref{eqn_pflj} and \cref{eqn_res_phot}.

The resulting Jacobian $\mathbf{H}^{pho}$ is calculated as:
\begin{equation}\label{eqn:jacobian}
{\mathbf{H}^{pho}_j} = \frac{\partial\mathcal{I}[{}_C\mathbf{p}_j]}{\partial {}_C\mathbf{p}_j} \cdot \frac{\partial \Pi({}_{L_j}\mathbf{p}_j)}{\partial {}_{L_j}\mathbf{p}_j} \cdot \frac{\partial {}_{L_j}\mathbf{p}_j}{\partial \Tilde{\mathbf{x}}}
\end{equation}
\vspace{-2mm}
\begin{equation}\label{eqn:proj_jacobian}\frac{\partial \Pi({}_{L_j}\mathbf{p}_j)}{\partial {}_{L_j}\mathbf{p}_j}  = \begin{bmatrix}\frac{-f_x y}{x^2+y^2}  & \frac{f_x x}{x^2+y^2} & 0 \\ -\frac{f_y x z}{LR^2}  & -\frac{f_y y z}{LR^2}& \frac{f_y L}{R^2}  \end{bmatrix}\end{equation}
\vspace{-2mm}
\begin{equation}
\resizebox{0.9\linewidth}{!}{
$\frac{\partial {}_{L_j}\mathbf{p}_j}{\partial \Tilde{\mathbf{x}}} = (\mathbf{R}_{L_jL_k} \mathbf{R}_{L_kI})\begin{bmatrix}[\mathbf{R}_{IG} ({}_G\mathbf{p}_j - {}_{G}\mathbf{p}_{GI})]_\times & -\mathbf{R}_{IG} & \mathbf{0}\end{bmatrix}$
}
\end{equation}
\resizebox{4em}{!}{
$\frac{\partial\mathcal{I}[{}_C\mathbf{p}_j]}{\partial {}_C\mathbf{p}_j}$ } is the image gradient from neighboring pixels.
We stack the point-to-plane (${}^{geo}$) and photometric (${}^{pho}$) terms into a combined residual vector ($\mathbf{z}$) and Jacobian ($\mathbf{H}$). The scaling factor $\sigma$ compensates for the different error magnitudes between geometric and photometric residuals:

$\mathbf{H}=\begin{bmatrix}{\mathbf{H}^{geo}_1}^T, \cdot \cdot \cdot , {\mathbf{H}^{geo}_m}^T, \lambda \cdot {\mathbf{H}^{pho}_1}^T,  \cdot \cdot \cdot , \lambda \cdot {\mathbf{H}^{pho}_n}^T\end{bmatrix}^T$
$\mathbf{z}_k^\kappa = \begin{bmatrix}{z^{geo}_1}, \cdot \cdot \cdot , {z^{geo}_m}, \lambda \cdot {z^{pho}_1},  \cdot \cdot \cdot , \lambda \cdot {z^{pho}_n}\end{bmatrix}^T$
$\mathbf{R}=\text{diag}\begin{bmatrix}\sigma\end{bmatrix} \vspace{1mm}$

We use the formulas provided in \cite{fastlio2} to update the state:
\vspace{-2mm}
\begin{equation}\mathbf{K}=(\mathbf{H}^T\mathbf{R}^{-1}\mathbf{H} + \mathbf{P}^{-1})^{-1}\mathbf{H}^T\mathbf{R}^{-1}\end{equation}
\vspace{-6mm}
\begin{equation} \resizebox{0.87\linewidth}{!} {$\widehat{\mathbf{x}}_k^{\kappa+1}=\widehat{\mathbf{x}}_k^\kappa \boxplus\left(-\mathbf{K} \mathbf{z}_k^\kappa-(\mathbf{I}-\mathbf{K H})\left(\mathbf{J}^\kappa\right)^{-1}\left(\widehat{\mathbf{x}}_k^\kappa \boxminus \widehat{\mathbf{x}}_k\right)\right)$ } \end{equation}
\vspace{-6mm}

\section{Experimental Results}
We quantitatively compare our proposed pipeline with several state-of-the-art approaches as baselines: KISS-ICP~\cite{kiss-icp}, LIO-SAM~\cite{lio-sam} and FAST-LIO2~\cite{fastlio2} represent widely used \ac{LO} and LIO algorithms. Similar to our approach, MD-SLAM~\cite{mdslam}, Du and Beltrame~\cite{realtimeint} and RI-LIO~\cite{ri-lio} also use intensity or reflectivity information. We use the Newer College Dataset~\cite{newercollege_ext} as a public baseline. To evaluate robustness in low-structured environments, we additionally provide and evaluate on a new dataset of geometrically degenerate scenes that is presented in Section \ref{dataset}. 
We calculated the absolute translational error (ATE) and the relative translational error (RTE) over segments of 10 m using the evo library\footnote{\url{https://github.com/MichaelGrupp/evo}}. We declare approaches with an RTE that is larger than $20\%$ as failed (indicated by $\times$) and do not report their ATE, as the required alignment between estimated and ground truth trajectories is not meaningful if the estimated trajectory differs too much from the ground truth.
Apart from sensor extrinsics, calibrations, and minimum range (to adapt for narrow scenes), we used the default parameters that were provided by the baseline approaches. We slightly increased the reflectivity covariance parameter in RI-LIO, as the default value caused divergence in all tested sequences.

\subsection{Newer College Results}
The Newer College Dataset~\cite{newercollege_ext} uses a hand-held 128-beam Ouster OS0. We present the results in Table \ref{tab:res_ncd}. In the \textit{Cloister} sequence, which contains large structures and slow motions, all approaches achieve a low ATE. In \textit{Quad-Hard}, aggressive rotations occur. Due to the absence of accurate ego-motion compensation, the LO approaches perform worst. Our approach achieves the lowest ATE, which confirms that our computationally cheap image motion-compensation method is effective. The \textit{Stairs} sequence causes most approaches to diverge. They use spatial downsampling of the point cloud to achieve real-time performance, which in the case of this narrow stairway removes too much information. While our approach uses the same downsampling for the geometric part, it achieves robust and accurate performance thanks to the photometric component. This unveils an inherent benefit of image-based intensity augmentation: fixed-size patches in the image implicitly capture different amounts of volume depending on the point distance. Thereby, projected images automatically have an adaptive resolution at a constant cost, contrary to the increased cost that would result from the required higher voxel resolution to capture the same information. 
While RI-LIO uses information from reflectivity images, its random feature selection fails to extract salient information and therefore diverges. In contrast, the dense approach in MD-SLAM does not fail but is outperformed by our approach. We perform slightly better than FAST-LIO2 on the longest and geometry-rich \textit{Park} dataset, showing that the intensity features can also improve performance in non-degenerate scenarios. 
We also evaluate our runtime on the \textit{Park} sequence. On average, our approach consumes 29.7 ms per frame (33 Hz) on an Intel i7-11800H mobile CPU, of which only 6.2 ms are spent on the photometric components, which shows that the main computational cost results from the conventional geometric approach.

\begin{table}[t!]
\caption{Newer College Dataset \\ \vspace{-1.5mm} Absolute Trajectory Error (RMSE) (\textit{m}) / Relative Error (\textit{\%})}
\vspace{-3mm}
\begin{adjustbox}{max width=\linewidth}
\begin{tabular}{ccccc}
\toprule
Method &  Quad-Hard & Cloister & Stairs & Park \\
Length \textit{(m)} & $234.81$ & $428.79$ & $57.04$ & $2396.20$ \\
\midrule
  KISS-ICP~\cite{kiss-icp} & $0.324$ / $1.88$ & $0.297$ / $2.07$  &  $\times$ / $32.48$ & $2.871$ / $1.06$\\
  MD-SLAM~\cite{mdslam} & $19.639$ / $12.36$ & $0.360$ / $2.73$ & $0.340$ / $6.21$ & $96.797$ / $23.03$ \\
  Du and Beltrame~\cite{realtimeint} & $18.506$ / $16.432$ & $59.544$ / $19.274$ & $\times$ / $26.121$ & $\times$ / $42.717$ \\
  LIO-SAM~\cite{lio-sam} & $0.299$ / $2.380$ & $0.145$ / $1.032$ & $\times$ / $5122.320$ & $1.566$ / $2.064$ \\
  FAST-LIO2~\cite{fastlio2} & $0.049$ / $\mathbf{0.26}$ & $\mathbf{0.078}$ / $\mathbf{0.23}$ & $\times$  / $3497.22$ & $0.310$ / $0.59$\\
  RI-LIO~\cite{ri-lio} & $0.237$ / $1.04$ & $0.285$ / $1.33$ &  $\times$ / $16877.28$ & $89.289$ / $5.00$  \\
  Ours & $\mathbf{0.046}$ / $0.29$ & $\mathbf{0.078}$ / $0.28$& $\mathbf{0.102}$ / $\mathbf{0.74}$ & $\mathbf{0.287}$ / $\mathbf{0.54}$ \\
\bottomrule
\end{tabular}
\end{adjustbox}
\label{tab:res_ncd}
\vspace{-7mm}
\end{table}

\subsection{ENWIDE Dataset} \label{dataset}
\begin{figure}[b!]
    \vspace{-5.5mm}
    \includegraphics[width=\linewidth]{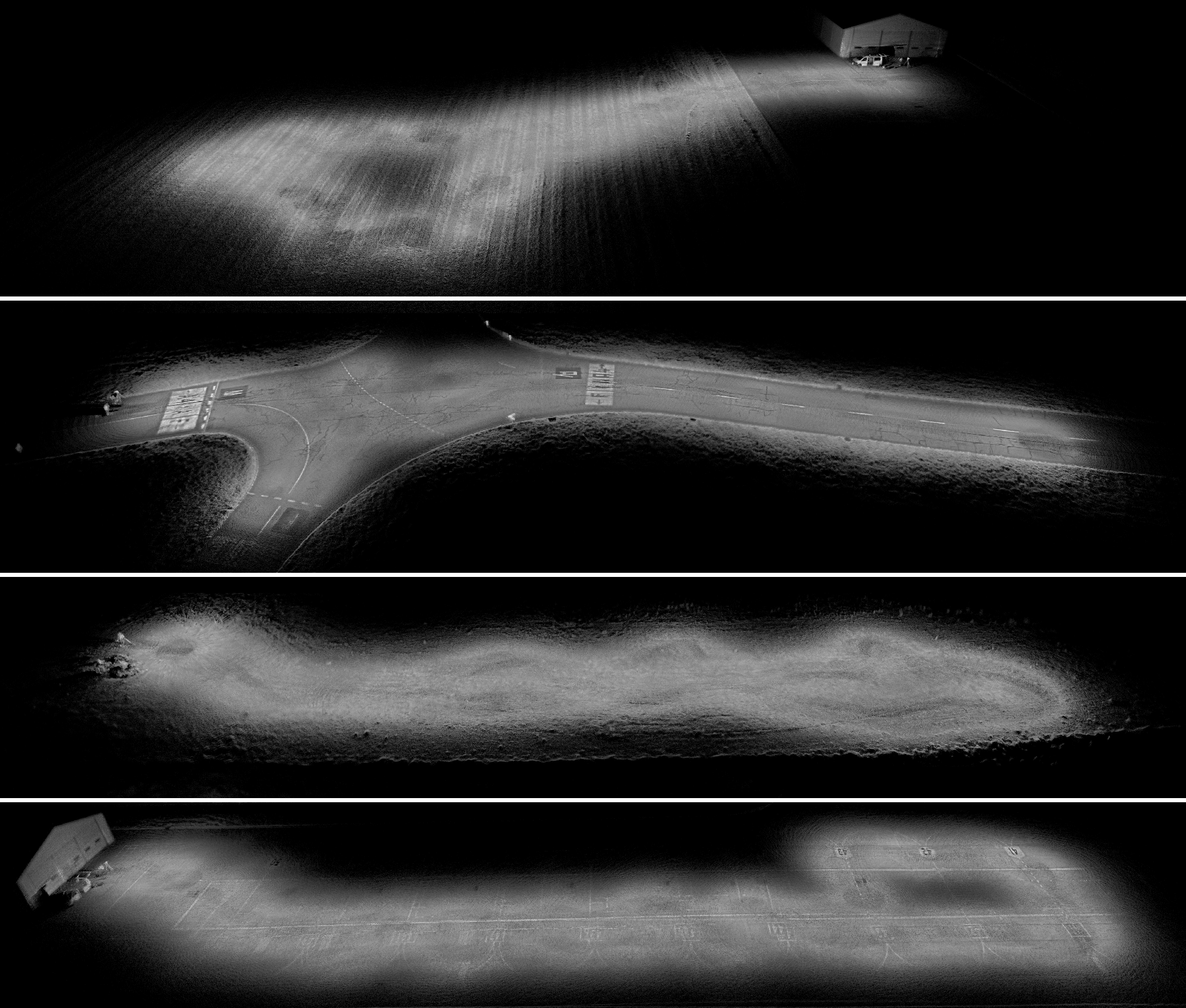}
    \vspace{-7mm}
    \caption{Resulting maps from COIN-LIO on the ENWIDE dataset. Top to bottom: FieldS, IntersectionS, KatzenseeS, RunwayS. Despite long degenerate sections, COIN-LIO produces consistent, sharp maps.}
    \label{fig:maps}
    \vspace{-1mm}
\end{figure}

As geometrically degenerate environments are barely represented in existing open-sourced datasets, we created a new dataset with long segments of real-world geometric degeneracy (Fig. \ref{fig:maps}). Using a hand-held Ouster OS0 128 beam LiDAR with integrated IMU, we recorded five distinct environments: \textit{Tunnel} (urban, indoor), \textit{Intersection} (urban, outdoor), \textit{Runway} (outdoor, urban), \textit{Field} (outdoor, nature), \textit{Katzensee} (outdoor, nature). All sequences contain long sections of geometric degeneracy but start and end in well-constrained areas. \textit{Tunnel/Intersection/Runway} sequences contain strong intensity features, \textit{Katzensee/Field} contain few salient features. For each environment, we provide one smooth (walking, slow motions) and one dynamic (running, aggressive motions) sequence. Ground truth positions were recorded from a Leica MS60 station with approximately 3 cm accuracy.

\subsection{ENWIDE Results}

\begin{table*}[t!]
\caption{\vspace{-1.5mm} ENWIDE Dataset - Absolute Trajectory Error (RMSE) (\textit{m}) / Relative Error (\textit{\%}) }
\vspace{-4mm}
\begin{adjustbox}{max width=\textwidth}
\begin{tabular}{ccccccccccc}
\toprule
Method &  TunnelS & TunnelD & IntersectionS & IntersectionD & RunwayS & RunwayD & FieldS & FieldD & KatzenseeS & KatzenseeD\\ 
Length \textit{(m)} & $251.58$ & $179.71$ & $279.28$ & $388.47$ & $333.57$ & $357.14$ & $232.70$ & $287.91$ & $242.88$ & $177.20$\\
\midrule
  KISS-ICP~\cite{kiss-icp} & $\times$ / $144.41$  & $\times$ / $68.11$ & $\times$ / $65.69$ & $\times$ / $64.84$ & $\times$ / $113.45$ & $\times$ / $124.64$ & $\times$ / $54.84$ & $\times$ / $70.70$ & $\times$ / $66.23$ & $\times$ / $76.80$\\
  MD-SLAM~\cite{mdslam} &  $\times$ / $88.16$ &  $\times$ / $80.76$ &  $\times$ / $90.87$ &  $\times$ / $87.89$ &  $\times$ / $97.73$ &  $\times$  / $91.13$&  $\times$ / $96.03$ &  $\times$ / $84.86$ &  $\times$ / $93.92$ &  $\times$ / $91.29$ \\
  Du and Beltrame~\cite{realtimeint} &  $\times$ / $58.084$ &  $\times$ / $56.086$ &  $\times$ / $60.812$ &  $\times$ / $57.449$ &  $\times$ / $67.906$ &  $\times$  / $63.978$&  $\times$ / $72.548$ &  $\times$ / $69.480$ &  $\times$ / $93.92$ &  $\times$ / $74.207$ \\
  LIO-SAM~\cite{lio-sam} &  $\times$ / $2565.621$ &  $\times$ / $2662.983$ &  $\times$ / $2022.878$ &  $\times$ / $2362.314$ &  $\times$ / $2334.174$ &  $\times$  / $3984.588$&  $\times$ / $2196.344$ &  $\times$ / $1999.968$ &  $5.588$ / $2.673$ &  $\times$ / $1485.377$ \\
  FAST-LIO2~\cite{fastlio2} & $\times$ / $316.12$  & $\times$ / $81.31$& $12.473$ / $29.28$  & $23.800$ / $28.11$ & $\times$ / $53.64$ & $\times$ / $59.84$ & $\mathbf{0.163}$ / $\mathbf{0.57}$ & $9.209$ / $16.08$ & $1.122$ / $4.31$& $1.02$ / $2.38$\\
  RI-LIO~\cite{ri-lio} & $\times$ / $70.32$ & $\times$ / $63.02$ & $\times$ / $49.94$& $\times$ / $188.83$& $\times$ / $52.18$& $\times$ / $79.16$ & $1.721$ / $2.44$ & $24.851$ / $25.89$& $\times$ / $49.34$& $\times$ / $154.19$\\
  Ours & $\mathbf{0.743}$ / $\mathbf{1.60}$ & $\mathbf{0.487}$ / $\mathbf{1.59}$ & $\mathbf{0.466}$ / $\mathbf{1.25}$ & $\mathbf{1.912}$ / $\mathbf{1.69}$  & $\mathbf{1.033}$ / $\mathbf{1.89}$ & $\mathbf{2.437}$ / $\mathbf{2.98}$  & $0.232$ / $0.85$ & $\mathbf{0.581}$ / $\mathbf{1.83}$ & $\mathbf{0.412}$ / $\mathbf{0.99}$ & $\mathbf{0.592}$ / $\mathbf{1.61}$ \\
\bottomrule
\end{tabular}
\end{adjustbox}
\label{tab:res_ours}
\vspace{-5mm}
\end{table*}
While our approach showed improved accuracy in Table \ref{tab:res_ncd}, the main motivation behind this work is to leverage intensity to improve robustness of LIO in challenging scenarios. We therefore evaluate on the challenging ENWIDE Dataset, presented in Table \ref{tab:res_ours}. It is plausible, that KISS-ICP fails in all sequences as it only operates on the (degenerate) point cloud geometry. However, we observe that MD-SLAM and Du, which also leverage the intensity channel, diverge in all sequences too. Both do not use the IMU, unlike LIO approaches, which impacts their ability to handle segments of geometric degeneracy or fast rotations. Additionally, Du only uses the images for geometric feature selection, and cannot benefit from additional texture information in the optimization.
Due to noise in the IMU measurements and drifting biases, LIO approaches can still fail in longer segments of geometric degeneracy. This is evident in LIO-SAM, which uses curvature-based point cloud features~\cite{loam}. FAST-LIO2, which operates on points directly, avoids divergence where the present vegetation still offers some weak information (\textit{Intersection}, \textit{Field}, \textit{Katzensee}) but exhibits large drift. However, we observe divergence in environments where the geometry is effectively perfectly degenerate (\textit{Tunnel}, \textit{Runway}). Despite using reflectivity, RI-LIO also diverges in most sequences. By leveraging the complementary information provided by the photometric error minimization, our approach achieves robust performance in all tested sequences,  While our main improvement is the increased robustness in difficult scenarios where other approaches fail, we also note higher accuracy than FAST-LIO2 on most successful sequences. 
\vspace{-1mm}

\subsection{Ablation study} \label{res:ablation}
\vspace{-1mm}
We show the effects of our image processing and feature selection in Table \ref{tab:ablation}.
We compare the proposed \textit{Filtered} image with \textit{Intensity} and \textit{Reflectivity} images. We also evaluate different feature selection policies by comparing \textit{Random} (similar to RI-LIO~\cite{ri-lio}) as well as \textit{Strongest} image gradient selection with the proposed geometrically \textit{Complementary} selection. 
We note lower error from (\textit{Intensity, Strongest}) than (\textit{Reflectivity, Strongest}). This seems surprising at first, as the reflectivity value compensates the range dependency of the signal. However, we observed that the reflectivity image contains stronger noise and artefacts and has less consistent brightness across the image. Our proposed image processing (\textit{Filtered, Strongest}) improves performance in low-textured environments (\textit{TunnelD, KatzenseeD}), where the line artefacts are more dominant than the actual features from the environment. Additionally, the brightness decreases drastically with increasing range. In contrast, the brightness compensation and line removal of our filtered image allow us to use more fine-grained details, e.g. from vegetation or gravel. We note that (\textit{Intensity, Strongest}) marginally outperforms (\textit{Filtered, Strongest}) on \textit{IntersectionS}. In this scene, strong image features from road cracks are consistently found at short range. We believe that the slightly lower ATE results from the filtered image having lower contrast than the intensity image at short range in this scene, which results in weaker gradients.
Selecting features based on strong image gradients (\textit{Filtered, Strong}) results in better performance compared to random patches (\textit{Filtered, Random}), as they provide richer information. Our proposed feature selection scheme (\textit{Filtered, Complementary}) achieves the best performance, as it reduces redundant information along uninformative geometric directions and specifically selects informative image patches. The impact is strongest in \textit{TunnelD}, where most gradients are in the geometrically degenerate direction along the tunnel (see Figure \ref{fig:complement}), while they are more randomly oriented in the other scenes. Overall, the ablation experiments confirm that COIN-LIO is able to effectively leverage the additional information provided by the multi-modality of the approach. %
\vspace{-0.5mm}

\begin{table}[t!]
\caption{\vspace{-1.5mm} Ablation Study - Absolute Trajectory Error (RMSE) (\textit{m})}
\vspace{-3mm}
\resizebox{\linewidth}{!}{
\centering
\begin{tabular}{cc|ccc}
\toprule
Image & Features &  TunnelD & IntersectionS & KatzenseeD \\ 
\midrule
Intensity & Strongest & $13.928$ & $0.472$ & $0.948$  \\
Reflectivity & Strongest &$\times$ & $0.699$ & $0.874$\\
Filtered & Strongest & $0.814$ & $0.489$ & $0.701$\\

Filtered & Random & $1.913$& $0.580$ &$1.073$\\

Filtered & Complementary  & $\mathbf{0.487}$ & $\mathbf{0.466}$ & $\mathbf{0.592}$\\
\bottomrule
\end{tabular}
}
\label{tab:ablation}
\vspace{-7mm}
\end{table}

\vspace{-1mm}
\section{Conclusion}
\vspace{-0.5mm}
We proposed COIN-LIO, a LiDAR-inertial odometry framework that fuses photometric error minimization on LiDAR intensity images with geometric registration to improve robustness in geometrically degenerate environments. We presented a filtering pipeline to produce brightness-compensated intensity images that provide more details and consistent illumination across different scenes. Our novel feature selection scheme effectively leverages the multi-modality by providing additional instead of redundant information. While COIN-LIO requires high-resolution LiDARs for dense intensity images, it slightly outperforms baseline approaches on the geometry-rich Newer College Dataset and shows drastically increased robustness in our new, geometrically degenerate ENWIDE dataset, which enables benchmarking in previously underrepresented scenarios.

We believe that this dataset as well as our work serve as a motivation for a new line of research that shifts from chasing even higher accuracy in geometrically simple cases to improving robustness in challenging environments. We also hope it motivates the industry to further improve the imaging capabilities of LiDAR.

\printbibliography
\end{document}